\documentclass[a4paper,twoside]{article}

\usepackage{epsfig}
\usepackage{subcaption}
\usepackage{calc}
\usepackage{amssymb}
\usepackage{amstext}
\usepackage{amsmath}
\usepackage{amsthm}
\usepackage{multicol}
\usepackage{xcolor}
\usepackage{pslatex}
\usepackage{apalike}
\usepackage{algorithm2e}
\usepackage[bottom]{footmisc}
\usepackage{hyperref}
\usepackage{SCITEPRESS}     % Please add other packages that you may need BEFORE the SCITEPRESS.sty package.

\begin{document}

\title{Calisthenics Skills Temporal Video Segmentation}%: A Videos and Body Keypoints Dataset}

\author{\authorname{Antonio Finocchiaro\sup{1}, Giovanni Maria Farinella\sup{1}\orcidAuthor{0000-0002-6034-0432}, Antonino Furnari\sup{1}\orcidAuthor{0000-0001-6911-0302}}
\affiliation{\sup{1}Department of Mathematics and Computer Science - University of Catania}
\email{1000006272@studium.unict.it,\{gfarinella, furnari\}@dmi.unict.it}
}


\abstract{Calisthenics is a fast-growing bodyweight discipline that consists of different categories, one of which is focused on skills. Skills in calisthenics encompass both static and dynamic elements performed by athletes. The evaluation of static skills is based on their difficulty level and the duration of the hold. Automated tools able to recognize isometric skills from a video by segmenting them to estimate their duration would be desirable to assist athletes in their training and judges during competitions. Although the video understanding literature on action recognition through body pose analysis is rich, no previous work has specifically addressed the problem of calisthenics skill temporal video segmentation. This study aims to provide an initial step towards the implementation of automated tools within the field of Calisthenics. To advance knowledge in this context, we propose a dataset of video footage of static calisthenics skills performed by athletes. Each video is annotated with a temporal segmentation which determines the extent of each skill. We hence report the results of a baseline approach to address the problem of skill temporal segmentation on the proposed dataset. The results highlight the feasibility of the proposed problem, while there is still room for improvement.
}

\onecolumn \maketitle \normalsize \setcounter{footnote}{0} \vfill

\section{\uppercase{Introduction}}
\label{sec:introduction}

The discipline of calisthenics is composed of various categories, including Skills, Endurance and Streetlifting\footnote{\url{https://en.wikipedia.org/wiki/Calisthenics}}. The skills category is the most popular and appreciated among those. It focuses on mastering challenging poses and movements that require high levels of strength, tendon stability, balance and coordination, engaging multiple upper and lower muscle groups simultaneously. 
In calisthenics, evaluation is generally performed by estimating the duration of a specific skill's hold.
Tools able to automatically segment a video in order to identify the execution of a skill and quantify its duration may be useful to support athletes in their training and judges during competitions.

\begin{figure}[t]
  \vspace{-0.2cm}
  \centering
  \includegraphics[width=\linewidth]{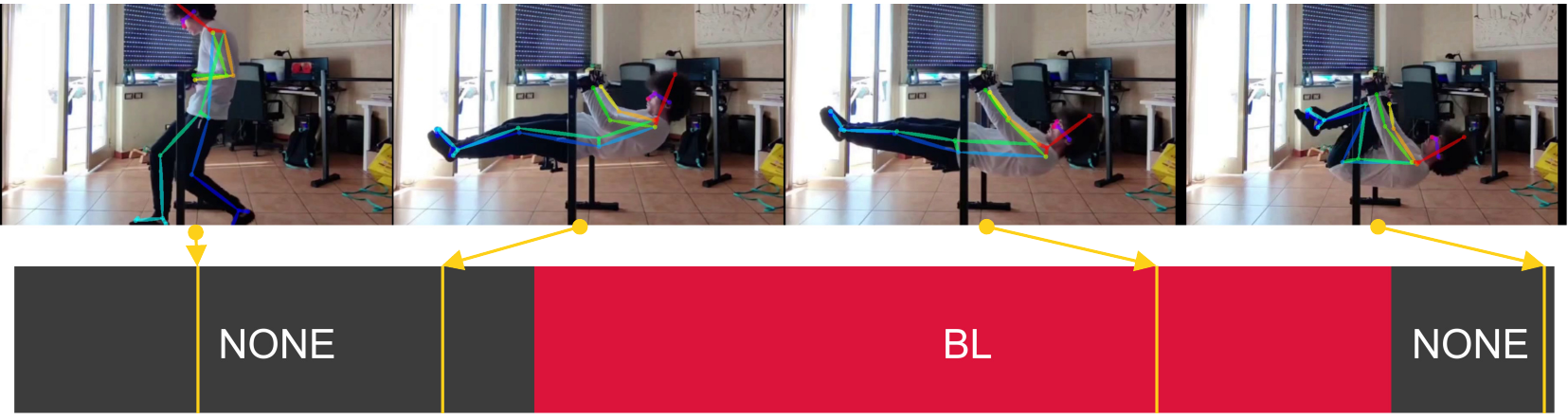}
  \caption{Calisthenics skill temporal video segmentation consists in breaking down a video into segments to highlight the beginning and end of each performed skill. As shown in the figure, the pose of the athlete has an important role in the considered task.}
  \label{fig:intro_fig}
\end{figure}

Although previous works have mainly focused on a range of team sports such as soccer~\cite{giancola2022}, basketball~\cite{khobdeh23}, tennis~\cite{Mora_2017_CVPR_Workshops} as well as individual sports such as swimming~\cite{s23042364}, badminton~\cite{rahmad20} or yoga~\cite{SURYAWANSHI2023109257} in this work, we consider the problem of calisthenics skills temporal video action segmentation (see Figure \ref{fig:intro_fig}), which can be used to provide assistance to calisthenics athletes during training or judges during competitions. This task has been accomplished by analyzing the 2D body pose of the athletes. In addition, previous works have not explicitly investigated algorithms for temporal segmentation of skills from video, which would be the core of such automated tools.
Aiming to provide an initial investigation on this topic, in this work, we contribute with a labeled dataset of videos of athletes performing calisthenics skills.
Specifically, we selected 9 skills based on their popularity among amateurs and professional athletes. The dataset contains $839$ videos of athletes performing skills, which have been collected from different sources including social networks and ad-hoc recordings, in order to ensure a realistic and natural set of video examples. 
%The main purpose of the designed dataset is to create a skills classification system which may assists athletes or judges in the evaluation phase by providing a temporal segmentation of the skills performed in footage. To achieve this goal, a specific pipeline has been designed. 
All videos have been manually labeled with temporal segmentation annotations which indicate the beginning and end of each skill execution.
We hence extracted the spatial coordinates of the athletes' joints in the videos using OpenPose \cite{cao2019openpose}. 
To provide initial results on the proposed dataset, we design a simple pipeline which includes a module performing a per-frame classification of body joint coordinates, followed by a module performing temporal reasoning on top of per-frame predictions.
The results highlight the feasibility of the proposed problem, while there is still room for improvement.
The dataset and code related to this paper are publicly available at the following URL: \url{https://github.com/fpv-iplab/calisthenics-skills-segmentation}.

\section{\uppercase{RELATED WORKS}}
This research is related to previous investigations in the fields of computer vision for sports analysis, human action recognition, and temporal video segmentation.

\subsection{Computer Vision for Sports Analysis}
In recent years, the application of computer vision based techniques for sports analysis has played a fundamental role in the development of automated tools capable of analyzing matches, providing statistics, or assisting the referees during competitions. 
Most of the previous works focused on team sports such as Soccer \cite{SPAGNOLO2014,MANAFIFARD201719,BANOTH22,app12157473,garnier2021evaluating}, Basketball \cite{Van_Zandycke_2022,info14010013,HAURI20,AHMADA20,8703164,ramanathan2016detecting}, Hockey \cite{koshkina2021contrastive} and many others \cite{martin2021automated}, \cite{9523123}, or individual disciplines such as Diving \cite{10097490}, Table Tennis \cite{kulkarni2021table} or Darts \cite{mcnally2021deepdarts}.
These works considered different image or video understanding tasks, including detecting and tracking objects and athletes \cite{liu2021detecting,9523003}. 
The reader is referred to~\cite{app12094429} for a review of video analysis in different sports.
Despite these advances, previous works did not consider the calisthenics field, hence resulting in a lack of datasets, tasks definitions and approaches. In this work, we aim to contribute an initial dataset and a baseline for the segmentation of calisthenics skills from video.

\subsection{Human Action Recognition}
Human Action Recognition (HAR) is a field of Computer Vision aiming to classify the actions performed by humans in a video. It consists of two main categories as discussed in \cite{YUE2022287,ren2020survey}:
\begin{itemize}    
    \item Skeleton-based recognition consists in studying the spatio-temporal correlations among various patterns of body joints. The spatial information provided by the joints can be extracted by a human pose estimation algorithm~\cite{9144178,10143178}. One possible implementation is related to graph convolutional networks as discussed in \cite{9552064}.
    \item RGB-based approaches use a different method to detect people, based on the analysis of RGB data within the images \cite{s21124246}. Although this approach can be trained end-to-end from videos, it generally needs to deal with the processing of irrelevant information (e.g., the background).
\end{itemize}
We observe that, in calisthenics skills video temporal segmentation, the athlete body pose plays an important role, while the background is less relevant. Hence, we base our analysis on the body joints extraction using OpenPose~\cite{cao2019openpose}.

% \subsubsection{Human Pose Estimation}
% Human Pose Estimation (HPE) is a field that aims to determine the pose of a human body by estimating the 2D or 3D spatial position of its joints \cite{9144178}. 
% It is based on two different approaches as discussed in \cite{10143178}.  
% The application of HPE is strongly related to concept of actions recognition. It is often considered one of the prerequisites for Human Action Recognition as discussed in \cite{SONG2021103055}.

\subsection{Temporal Video Segmentation}
Temporal Video Segmentation is an essential task in the field of video understanding~\cite{7780710,6909683,4813468}. It consists in dividing a video into relevant segments that represent the occurrence of predetermined events, such as human actions. The main task is to identify the boundaries of the events in the video in terms of timestamps or frames. In video action analysis, the application of this method allows us to analyze the actions performed by a human subject and their evolution over time. An introduction to this topic, the main adopted techniques and the most used evaluation metrics are discussed in \cite{ding2023temporal}. 
In this work, we consider the temporal video segmentation problem of segmenting calisthenics skill execution from video as a mean to estimate the starting, ending, and duration time of each skill. We base our experiments on the approach presented in~\cite{FURNARI20181}, which factorizes temporal video segmentation into per-frame processing and probabilistic temporal reasoning on top of per-frame predictions. We further compare this approach with a method based on heuristics for the reconstruction of the temporal segmentation of skills.
%One of the potential applications of Temporal Video Segmentation and a method to perform it, are presented in \cite{FURNARI20181} based on segmenting First Person Vision videos. 
%In this work, the presented method is used to discriminate skill segments and is compared to a experimental designed heuristic method.

\begin{figure*}[t]
  \vspace{-0.2cm}
  \centering
  \includegraphics[width=\linewidth]{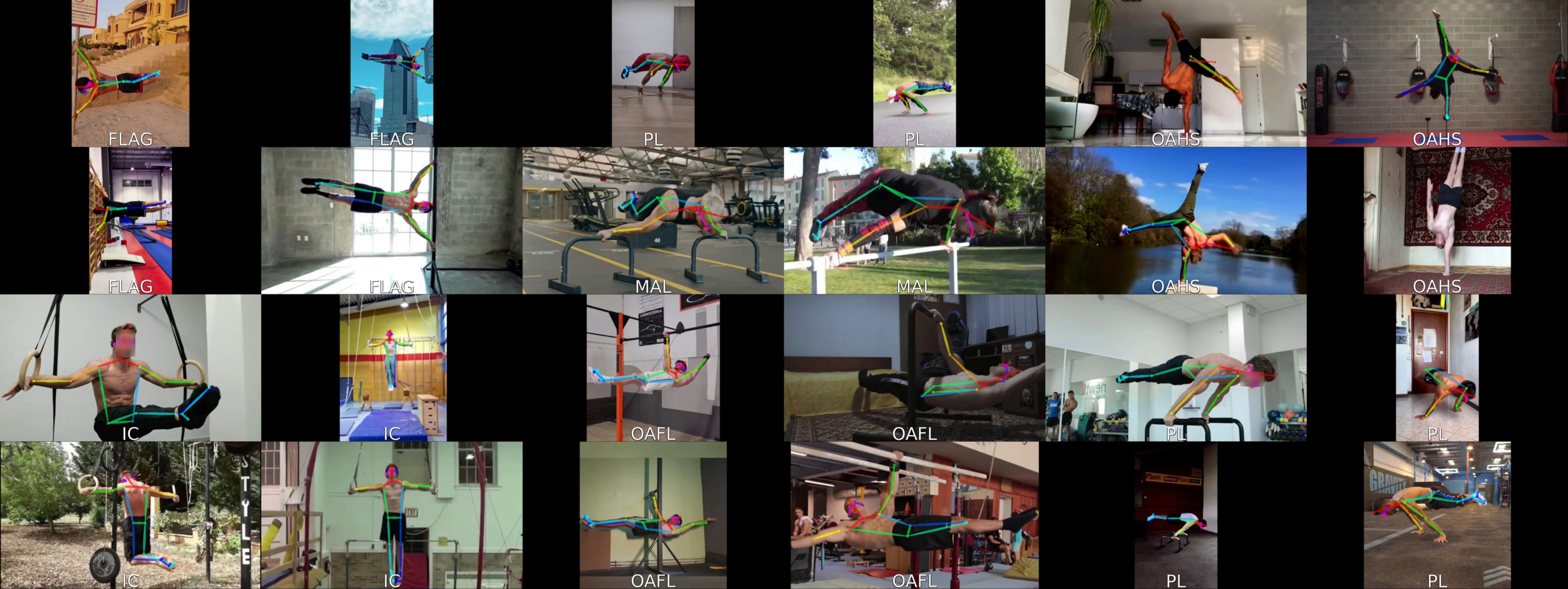}
  \caption{Example frames and related body poses from the proposed dataset.}
  \label{fig:collage_intro}
\end{figure*}

\section{\uppercase{Dataset}}\label{dataset}

In order to create a video dataset, we selected a set of skills to be covered. Each skill has a different difficulty level and demands a specific strength to be performed. All the chosen skills have been chosen based on their popularity among amateur and professional athletes and their importance in competitions such as the following: Burningate\footnote{\url{https://www.burningate.com/gare-calisthenics}}, SWUB\footnote{\url{https://streetworkoutultimatebattles.com}}, WOB\footnote{\url{https://worldofbarheroes.com}}, BOTB\footnote{\url{https://worldcalisthenics.org/battle-of-the-bars}}, WSWCF\footnote{\url{https://wswcf.org/competitions}}. 
 A detailed description of each skill is given in \cite{low2016}. The selected skills are listed in Table~\ref{tab:occurrences} together with the collection statistics.
% \begin{itemize}
%     \item Back Lever (bl)
%     \item Front Lever (fl)
%     \item Human Flag (flag)
%     \item Iron Cross (ic)
%     \item Maltese (mal)
%     \item One Arm Front Lever (oafl)
%     \item One Arm Handstand (oahs)
%     \item Planche (pl)
%     \item V-sit (vsit)
% \end{itemize}
We collected a total of 839 videos covering the 9 selected skills. Each video has been converted to a resolution of 960x540 pixels at 24 frames per second and trimmed so that each video comprises one skill. Each video can contain some `NONE' segments before or after the skill execution (see Figure~\ref{fig:intro_fig}) which models crucially have to recognize to quantify the actual duration of a skill. Almost all videos contain only the athlete in the scene, with few exceptions where multiple people are present in the background. In such cases, the athlete will still cover the foreground area. The most common backgrounds in the scenes include outdoor areas such as parks or streets, as well as in indoor locations like gyms or the athletes' homes. 
Figure~\ref{fig:collage_intro} shows some example frames from the proposed dataset. The average duration of skills in the videos is 5.83 seconds, the longest video has a duration of 27.83 seconds, whereas the shortest video lasts 0.83 seconds. The duration of the videos is strongly related to the complexity of the skill performed and the level of the athlete. 
For each collected video, we have manually labeled start/end times of the skill in frames and seconds, the corresponding skill category label, the MD5 checksum of the file and a video identifier composed of the skill name and a progressive number. 

\begin{table}
\caption{Occurrences of each skill in our dataset in terms of number of videos, seconds and frames.}
\label{tab:occurrences}
\resizebox{\columnwidth}{!}{%
\begin{tabular}{|l|l|l|l|}
\hline
\textbf{Skill}               & \textbf{Videos} & \textbf{Seconds} & \textbf{Frames} \\ \hline
Back Lever (BL)          & 88     & 574.66  & 13792  \\ \hline
Front Lever (FL)         & 108    & 443.08  & 10634  \\ \hline
Human Flag (FLAG)          & 75     & 633.16  & 15196  \\ \hline
Iron Cross (IC)          & 77     & 513.08  & 12314  \\ \hline
Maltese (MAL)             & 98     & 392.70  & 9425   \\ \hline
One Arm Front Lever (OAFL) & 80     & 363.50  & 8724   \\ \hline
One Arm Handstand (OAHS)   & 94     & 606.45  & 14555  \\ \hline
Planche (PL)             & 103    & 485.25  & 11646  \\ \hline
V-sit (VSIT)                & 116    & 814.87   & 19557 \\ \hline \hline
Total               & 839    & 4826.79 & 115843 \\ \hline
\end{tabular}
}
\end{table}

% \begin{figure}[t]
%   \vspace{-0.2cm}
%   \centering
%    {\epsfig{file = avg_durations.pdf, width=\linewidth}}
%   \caption{Histogram of the average duration of the videos in seconds.}
%   \label{fig:avg_durations}
% \end{figure}
 
To allow research on calisthenics skill recognition and temporal segmentation through body pose analysis, we extracted body poses from each frame through OpenPose~\cite{cao2019openpose}. We set the `number\_people\_max' flag to $1$ to allow the model to detect only the most prominent human when multiple subjects are present in the scene. This does not ensure that the recognized person is the athlete in a multi-person scene, so the video should contain only the athlete for optimal system performance. The `net\_resolution' parameter is set to $208$.

We considered the BODY\_25B model which can identify up to 25 human joints and provides three numerical values for each joint: the X coordinate, the Y coordinate and the confidence level. Body joint coordinates have been normalized by the frame dimensions in order to obtain values independent of the video resolution ranging from 0 to 1. As a result, each frame is associated to $75$ numerical features comprising $X$ and $Y$ coordinates of all body joints and related confidence scores.
%The output of the OpenPose processing consists of all the keypoints extracted frame by frame in a video, contained in $n$ JSON files (where $n$ = number of frames per video). Subsequently, a numerical dataset is generated starting from those values. In addition to the 75 columns representing the keypoints, this dataset includes the video name, frame number and a ground truth label representing the skills. The video segments outside of the skill section, are set to `none' movement. 

We divided the dataset randomly assigning $80\%$ of the videos to the training split and 20\% of the videos to the test split. 
%The split was performed at the video level to prevent overfitting at the frame level.  
The dataset and pre-extracted joints is available at \url{https://github.com/fpv-iplab/calisthenics-skills-segmentation}.

\section{\uppercase{METHOD}}
In this section, we describe the baseline calisthenics skills temporal segmentation approach used in our experiments. The proposed method is composed of two main modules: 1) a frame-based multiclass classifier which takes as input body joint features and predicts whether the current frame contains one of the skills or a 'NONE' background segment; 2) a temporal segmentation module which refines the per-frame predictions in order to obtain coherent temporal segments.
%dig into the various phases of the approach. We will begin with the creation of the dataset and some analysis related to it. Then, we will proceed to an instance of accomplishing a classification task.

\subsection{Multiclass Classifier}
This component is implemented as a Multilayer Perceptron. Its architecture consists of the first layer, which comprises 75 neurons, followed by three hidden layers, each composed of a linear layer and an activation layer. The chosen activation function is the LeakyReLU, which is shown to outperform other popular activation functions in the experiments. The output layer has $10$ nodes, corresponding to the $9$ skill classes, plus an additional `NONE' background class.
We train this module with a standard cross-entropy loss and Adam optimizer~\cite{kingma2017adam}. We use a batch size of 512 and train the model over 500 epochs, with the optimizer learning rate set to 0.0001. Optimal hyperparameter values were determined by cross-validation.

\subsection{Temporal Segmentation Module} \label{temp}
The temporal segmentation module works on top of the predictions of the multiclass classifier to produce coherent temporal segments.
The goal is to discriminate skill patterns and reconstruct the video timeline, trying to correct any mistaken skill predictions from the classifier. 
We consider two versions of this module: one based on a heuristic method and another one based on a probabilistic approach.
\subsubsection{Heuristic-Based Temporal Segmentation}
%All the frames that have been labelled by the classifier, are gathered into a video timeline to reconstruct the video. To achieve this, two temporal segmentation algorithms have been involved, and the benchmarks are illustrated afterwards. 
The heuristic approach aims to obtain coherent temporal segments from frame-wise predictions in three steps.
%The first algorithm involved is a heuristic method based on three steps. 

\noindent
\textbf{Sliding Window Mode Extractor (SWME)} This step relies on a sliding window process that iteratively returns the mode of the group of frames within the window. 
%WIP 
%

\begin{figure}[t]
  \vspace{-0.2cm}
  \centering
  \includegraphics[width=\linewidth]{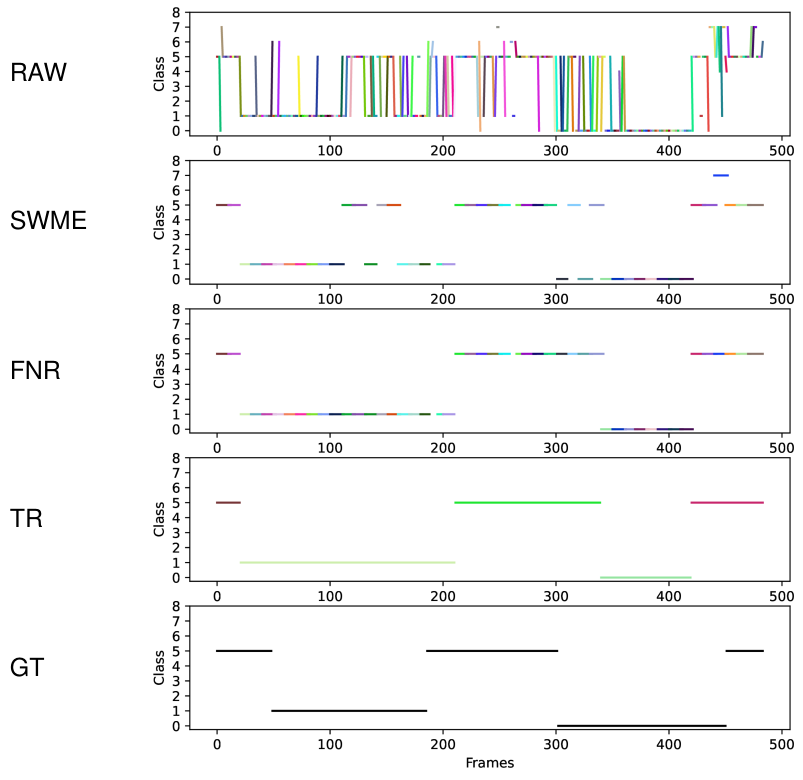}
  \caption{A representation of the heuristic algorithm applied to MLP predictions on a sample video. In the prediction of our algorithm, different colors represent different segments.} 
  %From top to bottom: Raw Predictions, Sliding Window Mode Extractor, Filtering Noise Removal, Timeline Reconstructor, Ground Truth.}
  \label{fig:vsr}
\end{figure}

Given the sequence of all $n$ frames present in the video, { $F = [f_0, f_1, \ldots, f_k, \ldots f_{n-1}]$,} the base window size is defined as:
\begin{equation*}
    w_b = \lfloor ((1-s) \cdot m \rfloor
\end{equation*}
{
with $m=32$ and $s$ defined as follows:
\begin{equation*}
s = 0.5 + \sum_{i = 0}^{n-2} (up [F_i = F_{i+1}] + dw [F_i \neq F_{i+1}])
\end{equation*}
where $up = \frac{0.14}{n}$, $dw = -\frac{0.11}{n}$.
The best-performing values for these constants have been defined following a naive approach of trial and error where $up$ and $dw$ represents respectively a reward and a penalty factor. The $m$ value consists of a multiplier factor which adjusts the window range from $[0.39,0.64]$ to $[11,19]$. Through these assignments, the size of the window is set as an inversely proportional ratio to the frame variance of the video. As can be seen, the divergences between contiguous frames are less weighted than equally labeled frames. This is related to the high frequency of different sequences of frame classes.}
%First, we define the size of the window. 
%To define the optimial size for the sliding window, we begin with a set that comprises a sequence of $n$ frames: $F = [f_1, f_2, \ldots, f_k, \ldots f_n]$ where $f_k$ represents the $k^{th}$ frame of the video.
%We define the following parameters: %Let %$n = |F|$, 
%$mult = 32$, $up = \frac{0.14}{n}$, $dw = -\frac{0.11}{n}$.
%We hence define $s \in \mathbb{R}$ as follows: 
% \begin{equation*}
%     s = \sum_{i = 0}^{n-2} (up [F_i = F_{i+1}] + dw [F_i \neq F_{i+1}])
% \end{equation*}
% Subsequently, the width of the $window$ is determined by: 
% \begin{equation*}
%     l_w = \lfloor ((1-s) \cdot mult \rfloor .
% \end{equation*}
We hence process the video with a sliding window of size $w_s$ which is initially set to $w_s=w_b$. Formally, the sliding window approach identifies subsets of $F$:
%This step requires the assignment of some variables and functions: 
%Let $stride = 3$ define the overlap between segments. The frame set is still denoted as $F$ and we initially set $ws = c = window$.
%Consider a subset $V \subset F$ with a cardinality of $|V|$ equal to $ws$. We iteratively define $V$ for each $c \in F$ as follows:
\begin{equation*}
    V^{(k)} = [f_{c-w_s+1}, f_{c-w_s+2},\ldots,  f_c] \>\>  \forall c \in \{0, n-1\}
\end{equation*}
The apex represents the $k^{th}$ subset iteratively taken in $F$.
Thus, we compute the mode of the considered subset. If more than one mode is found in $V^{(k)}$, there is no agreement in the current window, and we enlarge it by incrementing $w_s$ and $c$ by one unit (assuming that they do not exceed $n$). The mode-seeking process is hence iterated. When a single mode is found, $w_s$ is reset to $w_s=w_b$ and $c$ is incremented by $w_s-stride$ (we set $stride=3$) {to enlarge the window by one unit and calculate again the mode.}
%We calculate the frequency of each unique value: $f(x_i^{(v)}), \forall i \in V^{(v)}$.
%Find the maximum frequency: $f_{max} = \max(f(x_i^{(v)}))$.
%If there is only one unique value $x_m^{(v)}$ for which $f(x_m^{(v)}) = f_{max}^{(v)}$, $m$ is the unique mode. In this case, $ws$ is resetted to $window$, while $c$ is incremented by $ws-stride$ (assuring it does not exceed $n$), representing the last value of the new $V^{(v+1)}$ subset taken into account for mode calculation.
% Then, we define a frequency calculation function: \begin{equation*}
%     freq(x_i^{(v)}) = \sum_{j =0}^{m-1} [x_j^{(v)} = x_i^{(v)}] \forall i,j \in V
% \end{equation*}
% We define a variable to store the maximum frequencies: 
% \begin{equation*}
%     maxfreq^{(v)} = max_{i=0}^{m-1} (freq(x_i^{(v)}))
% \end{equation*}
% Next, we create a $Mode$ set that contains the modes in $V$.
% \begin{equation*}
%     Mode^{(v)} = \{x_i^{(v)} | freq(x_i^{(v)}) = maxfreq^{(v)}\}
% \end{equation*}
%If the mode is not unique, the variables $ws$ and $c$ are incremented by one (assuring they do not exceed $n$) and the mode is calculated again on the same subset $V^{(v)}$, expanded by the $c^{th}$ element. 
%The window size is iteratively calculated, updating the $ws$ value as follows: 
% \begin{equation*}
%     ws=\begin{cases}
% 			window,  & \text{if} |Mode|^{(v)} = 1\\
%             min(ws+1,m), & \text{otherwise}
% 		 \end{cases}
% \end{equation*}
For each step, we store the following three attributes about the local mode: 
\begin{equation*}
\begin{gathered}
idx_{start}(f_i)^{(k)} = min\{j | f_j^{(k)} = f_i^{(k)}\}\\
idx_{end}(f_i)^{(k)} = max\{j | f_j^{(k)} = f_i^{(k)}\}\\
mode^{(k)} = f_i^{(k)} \in Mode^{(k)} \Longleftrightarrow |Mode^{(k)}| = 1
\end{gathered}
\end{equation*}
These operations are repeated for each element in $F$, eventually resulting in the set:
\begin{equation*}
\begin{gathered}
W = \{I_0, I_1, I_2, \ldots, I_k, \ldots,I_{w-1}\}\\
\text{where } w = |W|\text{ and} \\
I_k = \{mode^{(k)}, idx_{start}(mode^{(k)})^{(k)}, idx_{end}(mode^{(k)})^{(k)}\} \\
\forall k \in F, h \in W
\end{gathered}
\end{equation*}
It is worth noting that we obtain $k \ll n$, where $n = |F|$. The output of this step consists of a smaller list of elements, each of which contains the three defined attributes. These are initial candidate video segments. Figure~\ref{fig:vsr} compares the results obtained in this step (second row) with the raw predictions (first row). As can be noted, this step significantly reduces the noise in the first row list, but may still contain incorrect segments due to high uncertainty.

\noindent
\textbf{Filtering and Noise Removal (FNR)} This step focuses on decreasing noise, applying a slightly modified version of the previous step.
Given the set $W$ from the previous step, this second stage returns a modified set, denoted as $R$: 
\begin{equation*}
    R = \{I'_0, I'_1, \ldots, I'_j, \ldots, I'_{w-1}\}  
\end{equation*}
Where the $j^{th}$ element is transformed as follows:

\begin{equation*}
\begin{aligned}
I'_j = & \text{mode}(I_{\max(j-2,0)}, I_{\max(j-1,0)}, I_j, \\
       & I_{\min(j+1,w-1)}, I_{\min(j+2,w-1)})
\end{aligned}
\end{equation*}
As observed in the third row in Figure \ref{fig:vsr}, some incorrect segments belonging to class 1, are effectively replaced with the local segment mode, providing a reliable reconstruction of the skill segment.
%It is important to note that this step is not parallelizable due to the fact that every elements is influenced by the $y$ preceding values, where $y \in [0,2]$.

\noindent
\textbf{Timeline Reconstructor (TR)} This step defines a new set denoted as $T$: 
\begin{equation*}
\begin{gathered}
    T = \{I''_0, I''_1, \ldots, I''_k, \ldots, I''_{t-1}\}  \\ \text{with }t < w
\end{gathered}
\end{equation*}
Where the $I''_k$ element is defined as follows: 
\begin{equation*}
I''_k=\begin{cases}
			I'_j,  & \text{if } x_j \neq x_z \> with \> x_j \in I'_j, x_z \in I'_{j+1}
                     \\
            merge(I'_j, I'_p), & \text{otherwise}
		 \end{cases}
   \end{equation*}
With the $merge$ function defined as follows: 
\begin{equation*}
\begin{gathered}
merge(I'_j, I'_p) = \{x_j, idx_{start}(x_j), idx_{end}(x_p)\} \\\forall j < p, j, p\in R
\end{gathered}
\end{equation*}
Through this step, different segments belonging to the same class are combined into a single segment representing a skill. The effect is illustrated in the fourth row in Figure \ref{fig:vsr}.
% More formally, let $(l_1^{(1)}, l_2^{(1)}, \ldots, l_n^{(1)})$ be the raw predictions obtained by the classifier. This processing step will output a list of refined predictions $(l_1^{(2)}, l_2^{(2)}, \ldots, l_n^{(2)})$, where
% \begin{equation}
% l_i^{(2)} = mode(l_{i-w_1/2}^{(1)},\ldots,l_i^{(1)},\ldots,l_{i+w_1/2}^{(1)})    
% \end{equation}
% and $w_1$ is the width of the sliding window.

%It is important to note that the predicted timeline is different from the reference one. This discrepancy arise from the classifier's incorrect predictions and is propagated through the pipeline.

\subsubsection{Probabilistic-Based Temporal Segmentation}
The probabilistic temporal segmentation module aims to output coherent temporal segments from per-frame predictions assuming a simple probabilistic model in which the probability of two consecutive frames having different classes is assumed to be low. We follow an approach similar to~\cite{FURNARI20181}.
%. It is based on a probabilistic method that relies on three stages: Discrimination, Negative Rejection and Sequential Modeling. The first two steps incorporate probabilistic concepts such as the posterior probability, while the third one employs HMM. 
%The method consists of grouping adjacent frames into video segments from the predictions of the classifier. It is called Negative Rejection and it is based on a Hidden Markov Model. This allows to obtain a higher accuracy, discouraging random labels transitions.
Let $F = \{f_1, \ldots, f_N\}$ be an input video with $N$ frames $f_i$ and let the corresponding set of labels be denoted as $L = \{y_1, \ldots, y_N\}$. 
The probability of labels $L$ is modeled assuming a Markovian model:
%By considering the conditional independence of the frames with respect to each other given the class and applying the Markovian assumption on the conditional probability distribution of class labels, the equation follows: 
\begin{equation*}
\begin{gathered}
P(L|F) \propto \prod_{i=2}^{n} P(y_i | y_{i-1}) \prod_{i=1}^{n} P(y_i | f_i).
\end{gathered}
\end{equation*}
Where the term $P(y_i | y_{i-1})$ represents the probability of transiting from a per-frame class to another. The transition probability is defined as follows: 
\begin{equation*}
    P(y_i | y_{i-1}) =
\begin{cases}
\epsilon, & \text{if } y_i \neq y_{i-1} \\
1 - M\epsilon, & \text{otherwise}
\end{cases}
\end{equation*}
The constant $\epsilon$ is hence defined as: $\epsilon \leq \frac{1}{M+1}$. Finally, the global set of optimal labels $L$ is obtained maximizing $P(L|F)$ using Viterbi algorithm: 
\begin{equation*}
    L = \arg\max_L P(L|F).
\end{equation*}

\section{EXPERIMENTAL SETTINGS AND RESULTS}
In this section we report experimental settings and results on the main components of the proposed pipeline.

\subsection{Multiclass Classifier}
% We performed an experimental analysis to determine the best parameters which maximize the performance of the Multilayer Perceptron component. Tests have been conducted using Cross-Validation with 5 folds. 
% Subsequently, the test results are examined through the confusion matrix and a frame level analysis. 
%
%\subsubsection{Network Benchmarks}
The optimal configuration for the Multilayer Perceptron (MLP) component has been determined through manual tuning of various hyperparameters.  

\noindent
\textbf{Optimizer}
%Figure \ref{fig:optim} displays the training loss trend of the different optimizers considered during a 500 epochs training. Adam and RMSProp have the fastest convergence, while RMSProp and SGD (with momentum = 0.9) achieved a slower decrease with higher convergence values. 
Four optimizers have been tested during a 500 epochs training. Among the chosen ones, Adam is the top performer, achieving 76.17\% accuracy in test phase and has the fastest convergence, closely followed by RMSProp that reaches a lowerT accuracy value equal to 74.72\%. In contrast, Adagrad and SGD (with momentum) delivery slower convergence and lower performance of 69.08\% and 68.61\% respectively. In the rest of the experiments, we use the Adam optimizer.

\noindent
\textbf{Activation Function}
Table \ref{tab:activf} compares the results when different activation functions are considered. For each function, the training loss, test accuracy, recall, precision and F1 score values are represented. LeakyReLU and ReLU have similar behavior with the first obtaining the best overall results. We use LeakyReLU in all subsequent experiments.

% \begin{figure}[t]
%   \vspace{-0.2cm}
%   \centering
%    {\epsfig{file = optimizer.pdf, width=\linewidth}}
%   \caption{Training Loss function trend with different optimizers.}
%   \label{fig:optim}
% \end{figure}

\begin{table*}[t]
\caption{Per-class ASF1 scores and related mASF1 measures for all compared methods.}
\label{tab:asf1}
\resizebox{\linewidth}{!}{%
\begin{tabular}{|l|l|l|l|l|l|l|l|l|l|l|l|}
\hline
Method    & mASF1 & BL    & FL    & FLAG  & IC    & MAL   & NONE  & OAFL  & OAHS  & PL    & VSIT  \\ \hline
Heuristic & 0.631 & 0.662 & 0.640 & 0.640 & 0.772 & 0.658 & 0.425 & 0.615 & 0.466 & 0.659 & 0.772 \\ \hline
Probabilistic &
  \textbf{0.674} &
  \textbf{0.723} &
  \textbf{0.681} &
  \textbf{0.641} &
  \textbf{0.847} &
  \textbf{0.713} &
  \textbf{0.476} &
  \textbf{0.658} &
  \textbf{0.492} &
  \textbf{0.714} &
  \textbf{0.797} \\ \hline
Raw       & 0.114 & 0.081 & 0.095 & 0.080 & 0.301 & 0.073 & 0.107 & 0.074 & 0.023 & 0.152 & 0.151 \\ \hline
\end{tabular}%
}
\end{table*}

\begin{table}[t]
\caption{Activation functions testing results comparison.}
\label{tab:activf}
\resizebox{\columnwidth}{!}{%
\begin{tabular}{|l|l|l|l|l|l|}
\hline
\begin{tabular}[c]{@{}l@{}}Activation \\ Function\end{tabular} & \begin{tabular}[c]{@{}l@{}}TR \\ Loss\end{tabular} & \begin{tabular}[c]{@{}l@{}}Test \\ Accuracy\end{tabular} & Recall         & Precision      & \begin{tabular}[c]{@{}l@{}}F1 \\ Score\end{tabular} \\ \hline
LeakyReLU                                                      & 0.008                                              & \textbf{76.17\%}                                         & 0.763          & \textbf{0.784} & \textbf{0.767}                                      \\ \hline
ReLU                                                           & 0.008                                              & 76.15\%                                                  & 0.762          & 0.782          & 0.765                                               \\ \hline
Sigmoid                                                        & 0.110                                              & 76.17\%                                                  & \textbf{0.765} & 0.768          & 0.764                                               \\ \hline
Tanh                                                           & \textbf{0.007}                                     & 74.49\%                                                  & 0.747          & 0.779          & 0.754                                               \\ \hline
SiLU                                                           & 0.029                                              & 74.69\%                                                  & 0.747          & 0.777          & 0.752                                               \\ \hline
\end{tabular}%
}
\end{table}

\begin{table}[t]
\caption{F1 Score per class.}
\label{tab:f1pc}
\resizebox{\linewidth}{!}{%
\begin{tabular}{|l|l|l|l|l|l|l|l|l|l|l|}
\hline
Skills                                              & BL   & FL   & FLAG & IC   & MAL  & NONE & OAFL & OAHS & PL   & VSIT \\ \hline
\begin{tabular}[c]{@{}l@{}}F1 \\ Score\end{tabular} & 0.83 & 0.74 & 0.83 & 0.91 & 0.74 & 0.63 & 0.76 & 0.64 & 0.80 & 0.88 \\ \hline
\end{tabular}%
}
\end{table}

%\subsubsection{Confusion Matrix Visual Analyses}
\noindent
\textbf{Per-class Results}
%The confusion matrix has been closely inspected trying to focus on the incorrect predicted classes. To this purpose, 
Table \ref{tab:f1pc} represents the F1 Score per-class. As can be seen, the Iron Cross has the highest value. This high level of accuracy is related to the fact that OpenPose can reliably detect poses for these skills, considering that Iron Cross is a vertical skill and it is not upside down, with full visible body (without limbs overlapping). One Arm Handstand, on the other hand, is the most challenging because of the upside down body position.  

\begin{figure}[t]
  \vspace{-0.2cm}
  \centering
  \includegraphics[width=\linewidth]{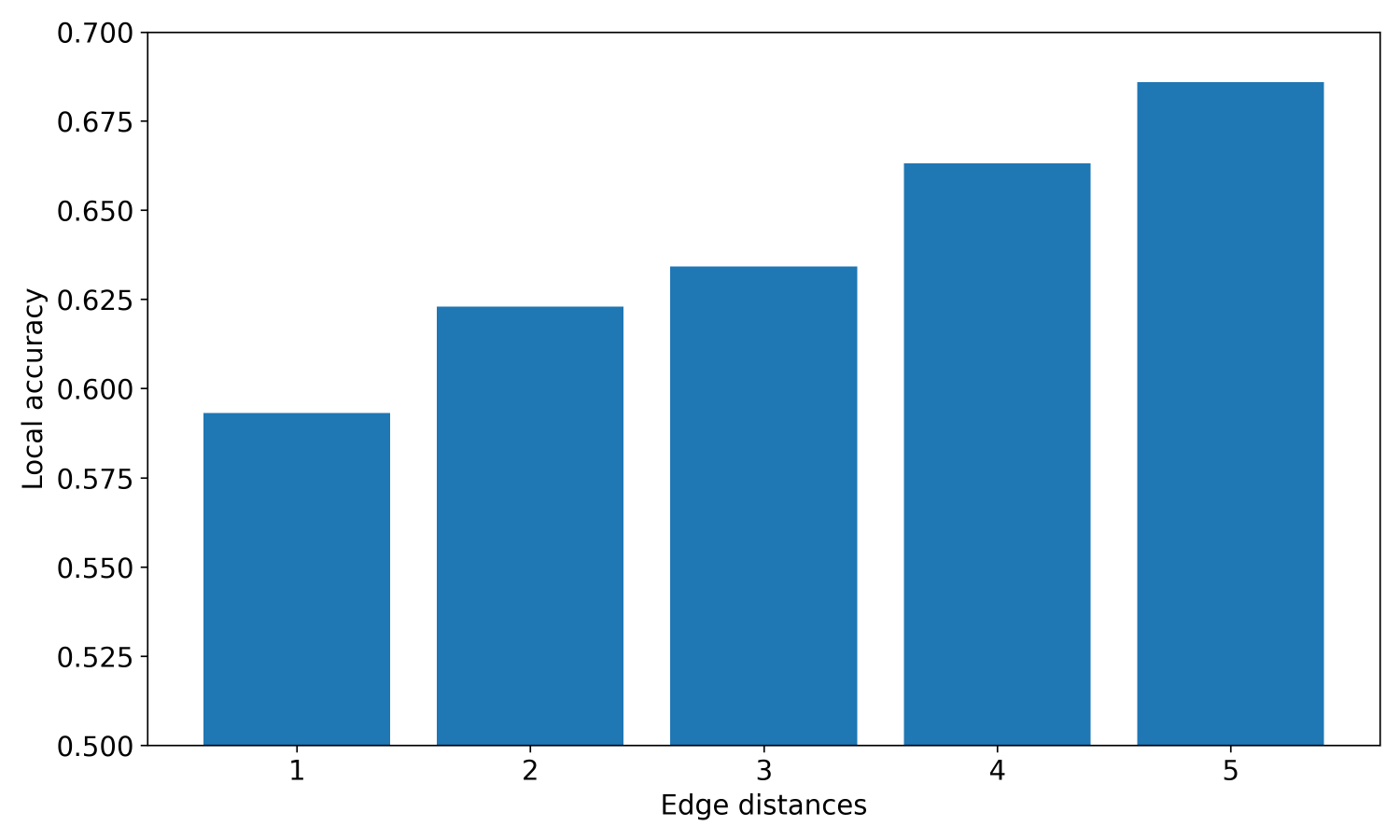}
  \caption{Accuracy (y axis) versus  distance of the classified frame from the edges of ground truth segments (x axis).}
  \label{fig:acc_edge}
\end{figure}

\noindent
\textbf{Accuracy at Segment Edges}
%\subsubsection{Frames Position and Accuracy Correlation}
We further investigated frame classification accuracy at different distances from the edges of ground truth skill segments.
The histogram shown in Figure \ref{fig:acc_edge} illustrates that frames closer to the edges of the segments generally have a lower accuracy than those with a more centered position (further from the edges). This correlation is related to the higher human and model uncertainty when classifying frames representing a transition from a skill to another movement or vice versa.

\subsection{Temporal Segmentation Algorithms}
%All the previous analyses were performed at frame level. The following analysis tests and compares the temporal segmentation algorithms described in \ref{temp}.
We evaluate the effectiveness of the temporal segmentation algorithms using the SF1 and ASF1 metrics considered in~\cite{FURNARI20181}.

The SF1 metric is a threshold-dependent, segment-based F1 measure. It is computed using precision and recall values for a specific threshold $\gamma$:
\begin{center}
\(
SF1^{(\gamma)}(t) = 2\cdot\frac{ \text{precision}^{(\gamma)}(t) \cdot \text{recall}^{(\gamma)}(t)}{\text{precision}^{(\gamma)}(t) + \text{recall}^{(\gamma)}(t)}
\)
\end{center} 
%The SF1 measure allows us to plot threshold-SF1 curves, which show the performance of the segmentation method at different tolerance levels. 
ASF1 (Average SF1) is the overall performance score of the segmentation method, computed as the average SF1 score across a set of thresholds \(T = \{t\text{ such that }0 \leq t \leq 1\}\):
\begin{center}
\(
ASF1^{(\gamma)} = \frac{\sum_{t \in T} \text{SF1}^{(\gamma)}(t)}{|T|}    
\)
\end{center}
mASF1 (mean ASF1) is the average ASF1 scores for all considered classes \((\gamma \in {0, ..., M})\). It provides an overall assessment of the method's performance across different classes.

\begin{figure}
  \vspace{-0.2cm}
  \centering
  \includegraphics[width=\linewidth]{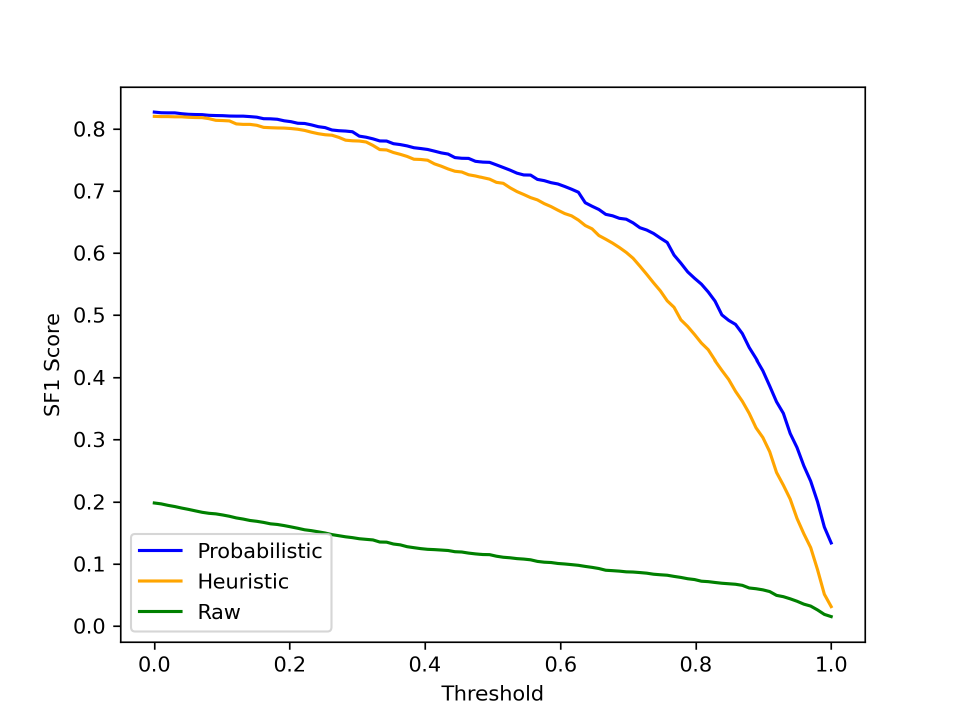}
  \caption{Threshold-SF1 curves comparing heuristic with respect to the probabilistic method.}
  \label{fig:sf1}
\end{figure}

%The evaluation was conducted during the test phase of the MLP, using its optimal configuration. 
Figure \ref{fig:sf1} illustrates the
SF1 scores of the algorithms at different thresholds $t$. Thresholds consist of 100 values comprised between 0 and 1 which influence the precision and the recall, hence the score.
As observed, the heuristic algorithm performs comparably across all thresholds, displaying a steeper decline in higher thresholds compared to the probabilistic one. In both analyses, the labels predicted by the MLP without the application of any temporal segmentation algorithm achieve limited performance. 
Table \ref{tab:asf1} provides the mASF1 and the ASF1 scores for each individual class.

\begin{figure*}
  \vspace{-0.2cm}
  \centering
  \includegraphics[width=\linewidth]{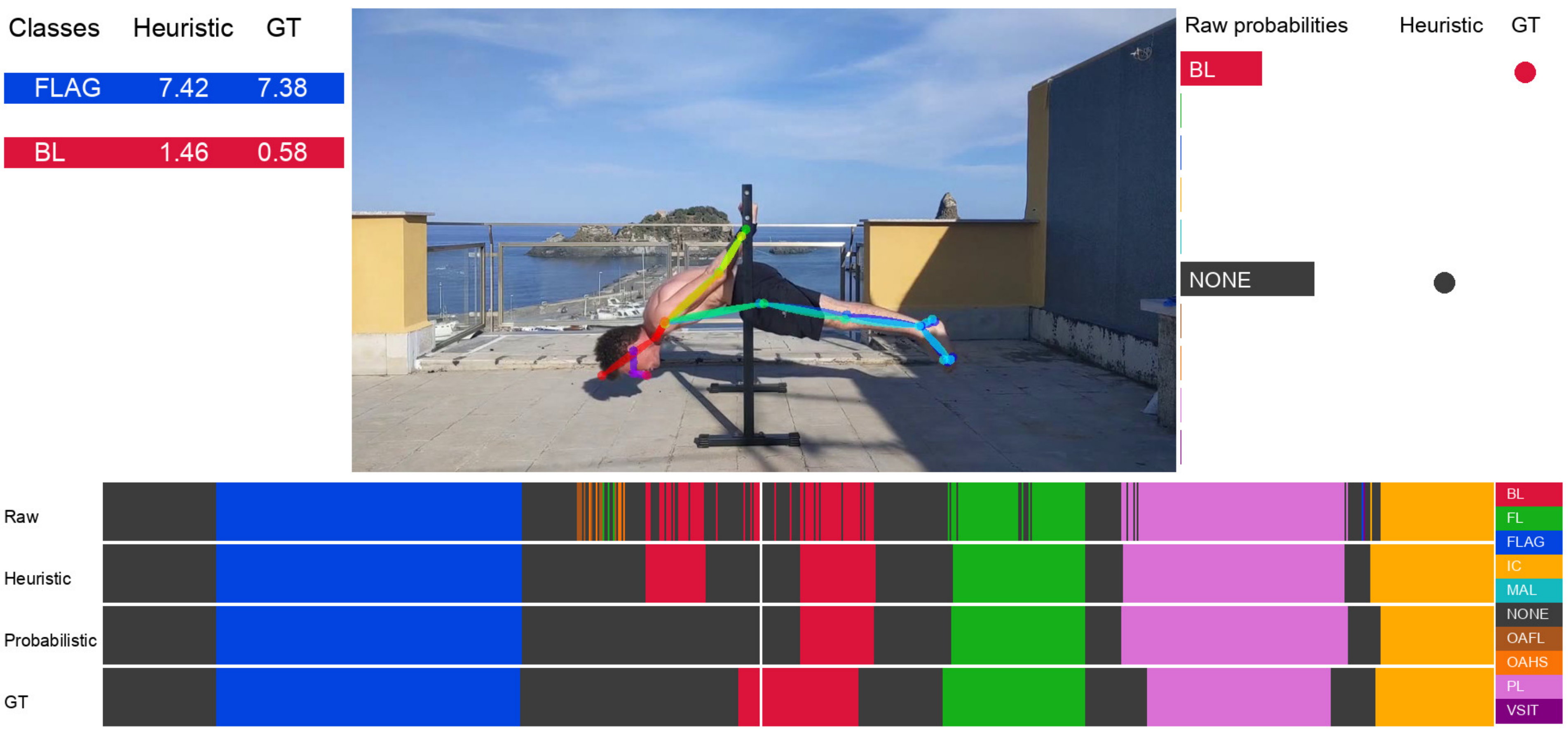}
  \caption{A scheme of the completed pipeline. From left to right: a list of the performed skills with their corresponding holding times, a frame showing the OpenPose tracking, a timeline consisting of four rows: raw prediction from the MLP, heuristic and probabilistic algorithm timeline reconstruction and the ground truth timeline. On the right, the output probabilities from the MLP and the class selected by the heuristic algorithm.}
  \label{fig:results}
\end{figure*}

The comparisons show that both algorithms achieve similar behavior in a range of situations included in the proposed dataset. We also note that while promising results are obtained, further enhancements can be made on the proposed dataset.
%Both algorithms were compared during the inference phase, showing similar behaviour when the MLP made accurate segment predictions. In other cases where MLP exhibited poor performance due to missing or wrong joints estimations, algorithms shows different behaviors. In general, the heuristic method, always determine segments, even in those where there is a relevant noise occlusion. This is explained by its mode calculation system. On other hand, the probabilistic method, has a proper HMM transition matrix system that rejects those segments with high variation. This analysis reveals situations in which the heuristic method identifies segments that are not present in the ground truth timeline. On other hand  the probabilistic method might not recognize ground truth elements.

An instance of the whole pipeline applied to a video is illustrated in Figure \ref{fig:results}. As can be observed, the set of skills present in the video is correctly identified. However, while the first, third, fourth and fifth segments are estimated with a great level of precision (small frame divergences do not affect the quality of holding time estimation), the second element (as displayed by the pose above), is not properly segmented. This issue is caused by incorrect predictions from the MLP, which cannot be rectified by the temporal segmentation algorithms. Improving the multiclass classifier could lead to better results.

\section{CONCLUSIONS}
In this work, we introduced a novel dataset of isometric calisthenics skills. To construct the dataset, we crawled videos and processed them using OpenPose, enabling us to retrieve the spatial coordinates of the body joints. We benchmarked a temporal segmentation approach based on per-frame classification performed with a Multilayer Perceptron and a temporal segmentation algorithm, for which we compare two versions, a heuristic-based approach and a probabilistic one.
Our analysis shows promising results, though there is potential for achieving better results through different architectures. We hope that the proposed dataset will support research in video analysis for Calisthenics skills recognition.

\section*{\uppercase{Acknowledgements}}
This research has been supported by Research Program PIAno di inCEntivi per la Ricerca di Ateneo 2020/2022 (C.d.A. del 29.04.2020) — Linea di Intervento 3 ``Starting Grant" - University of Catania.

\bibliographystyle{apalike}
{\small
\bibliography{main}}

\end{document}